\documentclass{article}
\usepackage{ijcai24}

\usepackage{times}
\usepackage{soul}
\usepackage{url}
\usepackage[hidelinks]{hyperref}
\usepackage[utf8]{inputenc}
\usepackage[small]{caption}
\usepackage{graphicx}
\usepackage{amsmath}
\usepackage{amsthm}
\usepackage{booktabs}
\usepackage{algorithm}
\usepackage{algorithmic}
\usepackage[switch]{lineno}

\usepackage{subcaption}
\usepackage{amsfonts}

\title{Improving Explainability of Softmax Classifiers Using a Prototype-Based Joint
Embedding Method}

\author{
Hilarie Sit$^{1,2}$\and
Brendan Keith$^3$\And
Karianne Bergen$^{1,2}$
\affiliations
$^1$Department of Earth, Environmental, and Planetary Sciences, Brown University, Providence, RI, USA \\
$^2$Data Science Institute, Brown University, Providence, RI, USA\\
$^3$Division of Applied Mathematics, Brown University, Providence, RI, USA\\
\emails
\{hilarie\_sit, brendan\_keith, karianne\_bergen\}@brown.edu
}

\begin{document}

\maketitle

\begin{abstract}
We propose a prototype-based approach for improving explainability of softmax classifiers that provides an understandable prediction confidence, generated through stochastic sampling of prototypes, and demonstrates potential for out of distribution detection (OOD). By modifying the model architecture and training to make predictions using similarities to any set of class examples from the training dataset, we acquire the ability to sample for prototypical examples that contributed to the prediction, which provide an instance-based explanation for the model's decision. Furthermore, by learning relationships between images from the training dataset through relative distances within the model's latent space, we obtain a metric for uncertainty that is better able to detect out of distribution data than softmax confidence.
\end{abstract}

\section{Introduction}
Deep neural networks have made their way into many scientific fields, achieving state-of-the-art performance on significant problems within these fields \cite{wang2023scientific}. However, the lack of transparency in their decision-making process along with their overconfident predictions reduce their trustworthiness in scientific applications where making an incorrect prediction can lead to severe consequences (\emph{e.g.,} medical diagnosis \cite{rudin2019stop}, extreme weather detection \cite{mcgovern2019making}). 

In this paper, we introduce an explainable artificial intelligence (XAI) method for retrieving explanations from a softmax classifier. We use an interpretable-by-design architecture in which the classification of a test image is decided solely from the similarities, as measured by distances in a shared latent space, between the image and representative examples of each class from the training dataset. This architecture closely follows that of prototype-based models, where the final portion of the neural network, commonly a softmax regression, makes an interpretable prediction using the latent space distances between the image and learned \emph{prototypes} (\emph{i.e.,} representations that have similarities to the training dataset in the feature space) \cite{li2018deep,chen2019looks}. One approach to obtain comprehensible explanations is to ensure that prototypes correspond to specific examples from the training dataset \cite{li2018deep}. Unlike standard prototype-based models with prototypes that are learned and then fixed during evaluation, our proposed model's prototypes can correspond to any combination of class examples, one image for each class, from the training dataset. 

Our approach may allow for a larger variety of prototypes that may be disregarded in other prototype-based methods. Also, the confidence obtained from stochastically sampling for prototypes can provide an uncertainty measurement that is calculated through samples unlike an arbitrary softmax confidence. % To achieve this flexibility, we combine the prototype-based architecture with a joint embedding method that uses a shared backbone network to map the class examples onto the same latent space as the image. Joint embedding methods are commonly used for self-supervised learning, in which relationships between images are learned by the model \cite{assran2023self,caron2021emerging,chen2020simple}. We use this method to learn the relationships, or distances in the latent space, between an image and a randomly sampled image from each class. 

An additional benefit of using distances for decision-making is that we gain a metric prior to the softmax layer that better captures epistemic uncertainty than softmax confidence, which lose this ability due to the feature overlap that occurs in the final layer of the network \cite{pearce2021understanding}. We demonstrate that these latent space distances better distinguish in-distribution datasets from out-of-distribution (OOD) datasets on OOD tasks. The contributions of this paper are as follows:
\begin{enumerate}
    \item Interpretable-by-design architecture that produces explanations for a softmax classifier's predictions
    \item A single forward pass uncertainty model that can be used for OOD detection
\end{enumerate}

\section{Related Works}
\textbf{Prototype-based Models.}
Prototype-based models are an example of an interpretable-by-design architecture that uses distances between an image's embedding and a collection of vectors in the latent space, corresponding to prototypes, in its prediction \cite{li2018deep}. Several works have also shown that parts or patches from the training images can be used as prototypes and demonstrate comparable performance to its standard neural network counterpart \cite{chen2019looks,nauta2023pip}. Prototype-based models are more transparent than standard neural networks by providing a visual explanation for how the model reached its decision, and variations of these models have been shown to be effective at out-of-distribution detection \cite{lu2024learning,sun2024classifier}.
\\ \\
\textbf{OOD Detection Methods.}
Single forward pass methods in the literature, such as DUQ \cite{van2020uncertainty}, SNGP \cite{liu2020simple} and DUE \cite{liu2020simple}, are more efficient and effective at OOD detection than traditional approaches for uncertainty estimation, such as deep ensembling \cite{lakshminarayanan2017simple} and Monte Carlo dropout \cite{gal2016dropout}. DUQ is most similar to our approach. While SNGP and DUE combine regularization with a Gaussian process to encode distance awareness, DUQ uses a radial basis function network that is trained using a two-sided gradient penalty to prevent feature collapse (a phenomenon where OOD data is mapped close to in-distribution data in the latent space) and preserve OOD detection performance, which can be measured as the distance to the closest centroid in the embedding space \cite{van2020uncertainty}.

\section{Prototype-Based Joint Embedding Method}
Our proposed Prototype-Based Joint embedding method (PB\&J) provides a framework for modifying a standard neural network architecture and training procedure to enable instance-based explanations through the use of a similarity-based terminal layer on the network.

\subsection{Dataset Generation}
A classification dataset consists of training instances (\emph{e.g.,} images) corresponding to $C$ different classes. Our method requires training pairs consisting of:
\begin{enumerate}
    \item $X$, a collection of $C+1$ randomly sampled instances containing an \textit{anchor}, $x_A \in \{S_1, S_2, \ \cdots \ ,S_C\}$, and $C$ \textit{class examples}, $x_i \in S_i \ \forall \ i \in \{1, 2, \ \cdots \ ,C \}$, where $S_i$ represents the subset of the training dataset that contains images from class $i$ 
    \item $y \in \{1, 2, \ \cdots \ ,C \}$, the classification label of $x_A$
\end{enumerate}

\subsection{Architecture}
The architecture starts with a standard feedforward neural network without its last fully connected layer. A trainable linear projection is used to project the flattened output from the network into a lower dimensional latent space, $\mathbb{R}^d$, where $d$ is a hyperparameter. This additional projection step has been shown by \cite{chen2020simple} to help address the dimensionality collapse problem, a type of partial collapse that occurs when the latent space vectors do not span the entire dimension of the latent space, in joint embedding methods. Weights in the network and projection layer are shared among all elements of $X$. From a forward pass through this backbone network $f_w$, we get latent space representations $\ell_A = f_w(x_A)$ for the anchor and $\ell_i = f_w(x_i) \ \forall \ i \in \{1, 2, \ \cdots \ ,C \}$ for the class examples.

To force the model to make predictions based on distances between these latent space representations, we use a similarity-based layer prior to the softmax. Here, a distance array $\bar{m} \in \mathbb{R}^C$ is computed for the anchor:
\begin{equation}
m_{i} = \log\bigg(\frac{d_i^2+1}{d_i^2+1e^{-10}}\bigg)
\label{eq:distance}
\end{equation}

where $d_i = ||\ell_A - \ell_i||_2$ is the Euclidean distance between the anchor $\ell_A$ to every class example $\ell_i$ in the latent space. This distance array is then multiplied with a matrix of trainable weights $W \in \mathbb{R}^{C \times C}$ to generate the score array $\bar{s} = \bar{m} \cdot W^T \in \mathbb{R}^C$. Probabilities for the anchor belonging in each class can be calculated by applying softmax to the score array. The last portion of the model is equivalent to applying softmax regression without biases on the distance array. The architecture is shown in Figure 1.

\begin{figure}[t]
  \centering
   \includegraphics[width=\linewidth]{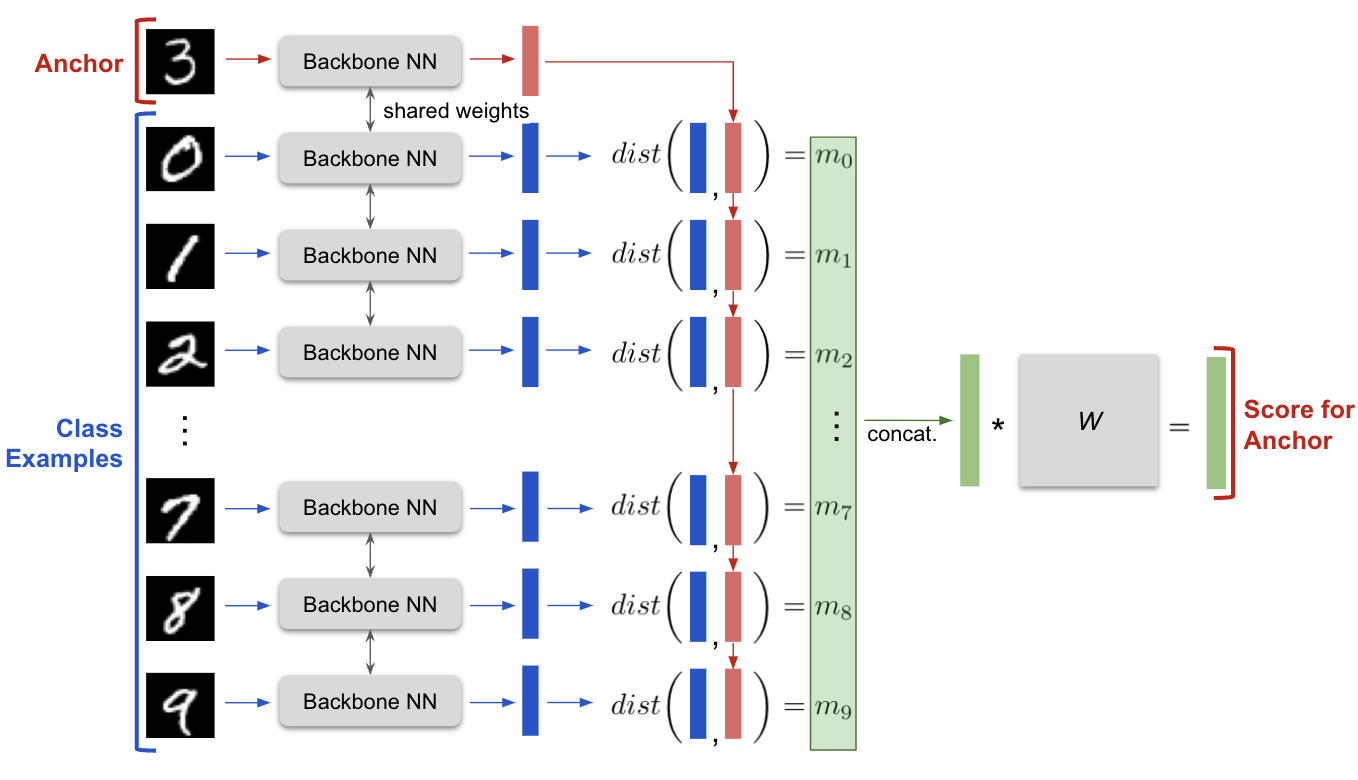}
   \caption{Model Architecture.}
   \label{fig:architecture}
\end{figure}

\subsection{Training}
The model is trained by minimizing the cross entropy loss on the classification label of the anchor, $y$.

%\subsubsection{Initialization}
\vspace{1em}
\noindent
\textbf{Initialization.} 
Weights within the backbone neural network $\theta_{BB}$ are randomly initialized and the weights within the weight matrix prior to the softmax layer, $W$, are initialized with a large positive value $\gamma$ on the diagonals and $-\gamma/(C-1)$ on the off-diagonals. This initialization scheme encourages the anchor to move toward examples of its own class in the latent space.

\vspace{1em}
\noindent
\textbf{Sampling for Class Examples.} We use an efficient sampling technique to obtain a set of class examples for training by sampling these from within the mini-batch. If the mini-batch does not include an instance from every class, we randomly sample one example from each missing class. In every update, a mini-batch of examples is fed into the backbone network to obtain latent space representations for the anchors. Because class examples are randomly chosen from within this set, their latent space representations do not need to be recalculated, which makes training time comparable to that of a standard neural network.

\subsection{Evaluation}
In PB\&J, there are two approaches to obtain predictions for unseen examples: (1) a stochastic sampling approach in which class examples are randomly sampled from the training dataset, and (2) a centroid-based approach in which class examples are replaced by centroids for each class in the latent space, calculated from all examples in the training dataset. The former approach can be used to retrieve explanations for the model's decision while the latter, less informative, approach can be used for single forward pass OOD detection.

\vspace{1em}
\noindent
\textbf{Stochastic Sampling.} \label{sec:stochastic} In the stochastic sampling approach, we randomly sample for a set of class examples from the training dataset and compute a prediction based on the latent space distances of the test image to those examples using our model. Here, we have access to the class examples that correspond to the prototypes that directly led to the prediction.

\vspace{1em}
\noindent
\textbf{Centroid-Based.}
In the centroid-based approach, class centroids are used in place of class examples to compute the distance array in Equation 1. These centroids are pre-computed from all of the training examples, which reduces computation time during evaluation. Additionally, evaluation only requires a single forward pass, which makes this approach more suitable for OOD detection. The OOD metric is the log distance between the unseen example to the closest centroid in latent space, $m$. The higher this value, the more likely it is the example belongs within the training distribution (\emph{i.e.,} in distribution). 

\section{Experiments}
We demonstrate our method's capabilities on several image classification datasets, including MNIST, FashionMNIST, CIFAR10, and CUB-200-2001. Architecture and training details can be found in the appendix. We report the classification accuracy in Tables 1 and 2. On MNIST, FashionMNIST, and CIFAR10, PB\&J's accuracy is comparable to its standard neural network counterpart. On CUB-200-2001, PBJ with a ResNet34 backbone achieves a test accuracy of $79.9\% \pm 0.2$, which is comparable to ProtoPNet's accuracy of $79.2\%$ \cite{chen2019looks}, and PB\&J with a ResNet50 backbone achieves a test accuracy of $83.8\% \pm 0.4$, which is higher than PIP-Net C's accuracy of $82.0\% \pm 0.3$ but lower than PIP-Net R's accuracy of $84.3\% \pm 0.2$ \cite{nauta2023pip}.

\begin{table}[t]
  \centering
  \begin{tabular}{@{}lrrr@{}}
    \toprule
    Dataset &  \shortstack{PB\&J \\ (Stochastic)} & \shortstack{PB\&J \\ (Centroid)} & \shortstack{Neural \\ Network}\\
    \midrule
    MNIST & $99.5$ & $99.5$ & $99.5$ \\
    FashionMNIST & $92.8$ & $92.8$ & $92.4$ \\
    CIFAR10 & $95.0$ & $95.0$ & $95.2$ \\
    % CUB-200-2001 & - & $79.9$ & $82.3$ \\
    \bottomrule
  \end{tabular}
  \caption{Classification accuracy on several image classification benchmarks.}
  \label{tab:accuracy}
\end{table}

\begin{table}[t]
  \centering
  \begin{tabular}{@{}lrrr@{}}
    \toprule
    Architecture & PB\&J & ProtoPNet & PIP-Net C/R\\
    \midrule
    ResNet32 & $79.9$ & $79.2$ & - \\
    ResNet50 & $83.8$ & - & $82.0/84.3$\\
    \bottomrule
  \end{tabular}
  \caption{Classification accuracy on CUB-200-2001.}
  \label{tab:accuracy}
\end{table}

\subsection{Explainability with Prototypes}
We applied the stochastic sampling approach from Section 3.4 on several challenging test images (\emph{i.e.,} images that were misclassified using the centroid-based approach) to gain insight into the model's decision-making process. For each of these images, $100$ sets of class examples (prototypes) are randomly sampled to make $100$ separate predictions. From these predictions, we plot a predictive distribution showing the probability of each class being predicted by the model.

For a misclassified \textit{coat} from FashionMNIST (Figure 2), the model is largely undecided between \textit{coat} and \textit{shirt}, as shown in Figure 2a. We can visualize which specific prototypes resulted in the conflicting predictions. For separate instances where the model predicted \textit{coat} and where the model predicted \textit{shirt}, we visualize the prototypes and their similarity score to the test image that led to each decision in Figure 2b. Similarity scores correspond to the log distances in latent space. The scores strongly contribute to the decision \--- in instances where the test image is closer (\emph{i.e.}, higher log distance) to the coat prototype than the shirt prototype, the model tends to predict \textit{coat} over \textit{shirt}, and vice versa.

\begin{figure}[t]
\centering
  \begin{subfigure}{0.25\linewidth}
\includegraphics[width=\linewidth]{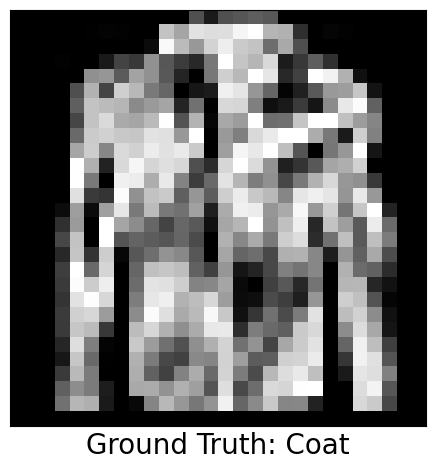}
    \caption{Test image}
    \label{fig:fmnist_image}
  \end{subfigure}
  \begin{subfigure}{0.74\linewidth}
\includegraphics[width=\linewidth]{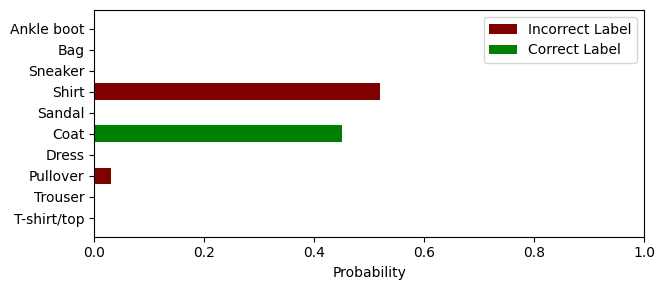}
    \caption{Predictive distribution}
    \label{fig:fmnist_posterior}
  \end{subfigure}
  
  \hfill
  
  \begin{subfigure}{\linewidth}
\includegraphics[width=\linewidth]{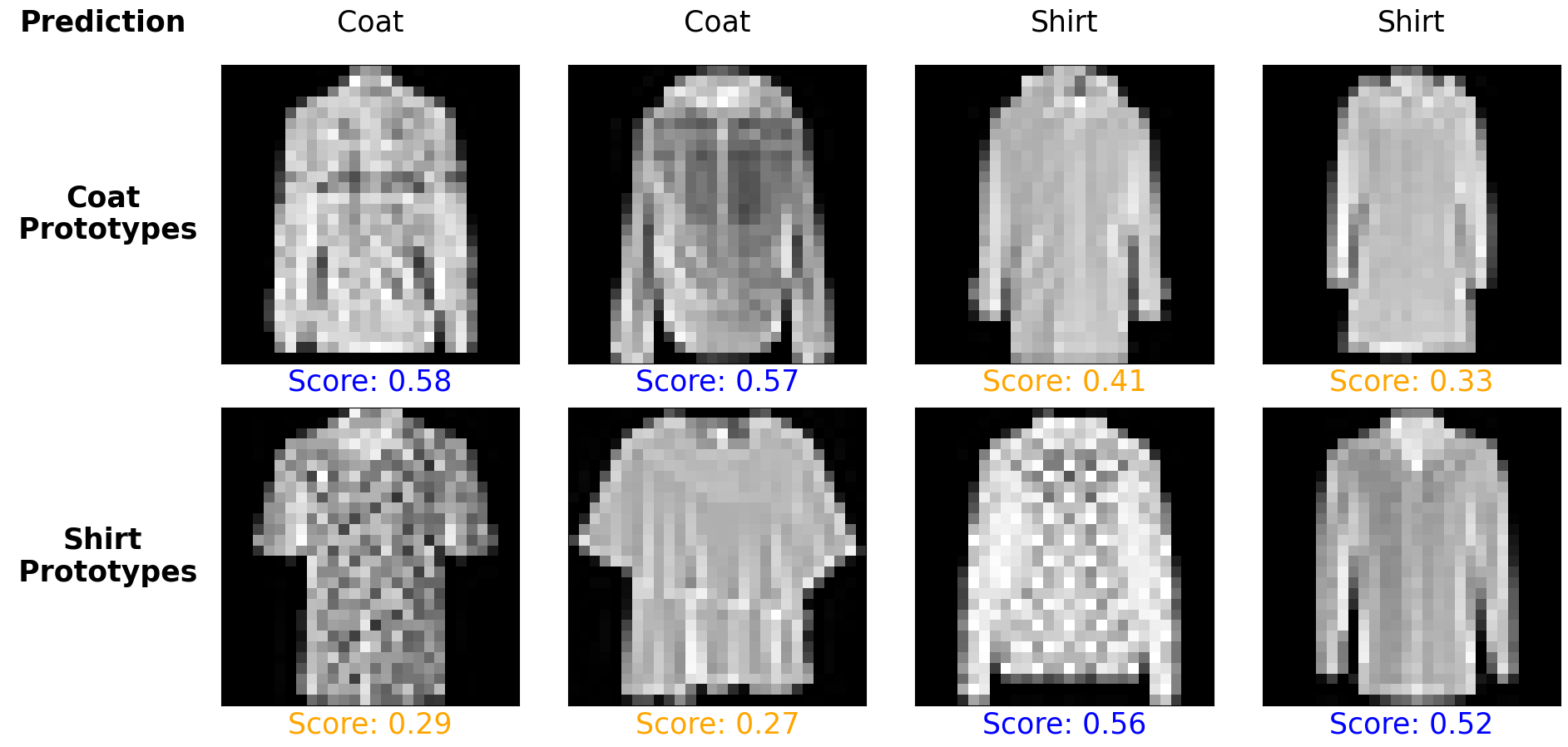}
    \caption{Closest prototype (blue) from four separate prediction instances and corresponding image from the other class (orange) in set of class examples}
    \label{fig:fmnist_prototypes}
  \end{subfigure}
  \caption{Analysis for a challenging image of a \textit{coat} from FashionMNIST. Model is undecided between \textit{coat} and \textit{shirt}, and we show the closest prototypes that resulted in each type of prediction.}
  \label{fig:fmnist}
\end{figure}

When the stochastic sampling approach is applied to less challenging and straightforward test images like Figure 3a, the model predicts the correct class $100\%$ of the time, as shown in Figure 3b. We observe that variance in the posterior distribution is correlated to similarities of images across different classes, which can be associated with aleatoric uncertainty. More examples are shown in the appendix.

\begin{figure}[t]
\centering
\begin{subfigure}[t]{0.25\linewidth}
\includegraphics[width=\linewidth]{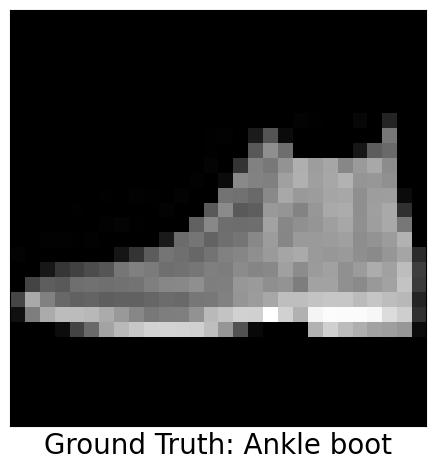}
\caption{Test image}
\label{fig:image_correct}
\end{subfigure}
\begin{subfigure}[t]{0.74\linewidth}
\includegraphics[width=\linewidth]{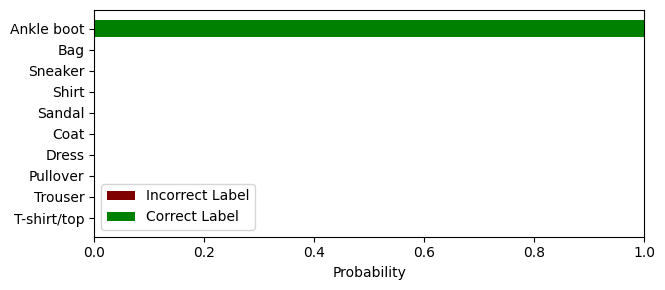}
\caption{Predictive distribution}
\label{fig:posterior_correct}
\end{subfigure}

  \hfill
  
  \begin{subfigure}{\linewidth}
\includegraphics[width=\linewidth]{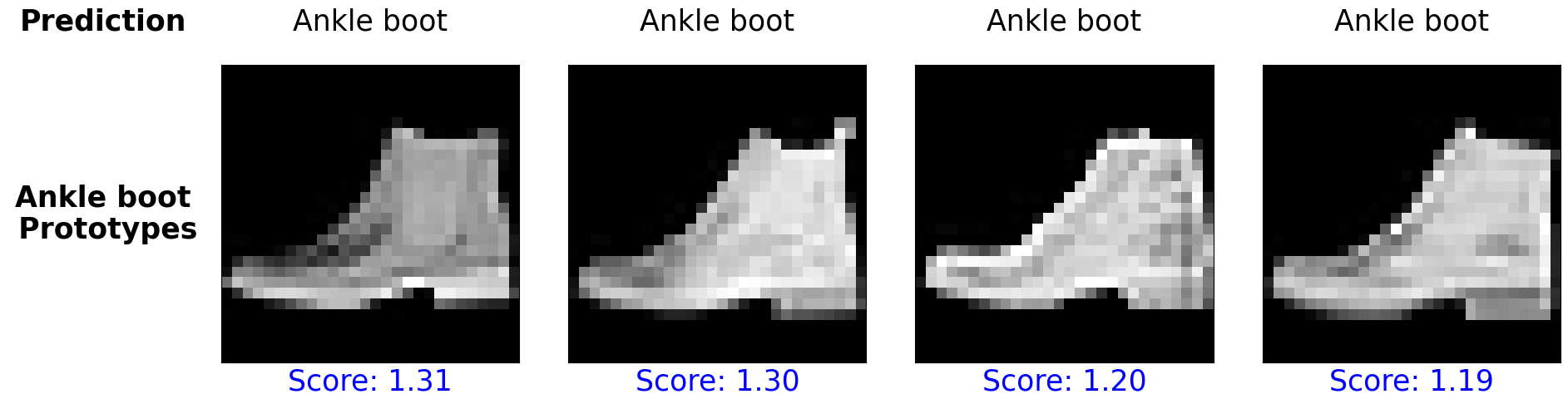}
    \caption{Closest prototype from four separate prediction instances}
    \label{fig:prototypes_correct}
  \end{subfigure}
  \caption{Analysis for a straightforward image of an \textit{ankle boot} from FashionMNIST. Model predicts the correct class 100\% of the time, and the closest prototypes are visually similar to the test image.}
  \label{fig:fmnist_correct}
\end{figure}

\vspace{1em}
\noindent
\textbf{Prediction and Confidence.}
From the stochastic sampling approach, we obtain an overall prediction, which corresponds to the most commonly predicted class, with a confidence, which is the percentage of instances where the model made that prediction. Unlike softmax confidence, our confidence is understandable, since it represents the occurrences of images from different classes that are most similar to the test set in the particular set of chosen class images.

% In Figure 4, for the first five misclassified test images in the test set, we plot the predictive distribution from the stochastic sampling approach along with softmax confidence from a standard neural network.

% \begin{figure}[t]
% \centering
% \includegraphics[width=\linewidth]{fmnist_softmax.png}
% \caption{Confidence from stochastic sampling approach compared to softmax confidence for first five misclassified images.}
% \label{fig:fmnist_softmax}
% \end{figure}

\subsection{OOD Detection Performance}
We perform a preliminary examination into our model's OOD detection potential with \texttt{scikit-learn}'s Two Moons dataset, MNIST vs FashionMNIST, and FashionMNIST vs MNIST.

The Two Moons dataset was used to visualize the method's OOD detection performance. We plot the uncertainty for a standard neural network (softmax confidence) and PB\&J (log distance from centroid) in Figure 4. To scale the log distances to a standardized metric, we will define the confidence of an example belonging in distribution as:

\begin{equation}
P_{ID} = \text{sigmoid}\bigg(\frac{m - \alpha}{\sigma^2}\bigg)
\end{equation}
where $\alpha$ represents the distance that encompasses $95\%$ percentile of the training dataset, and $\sigma^2$ is the standard deviation of the distances from the training dataset. These parameters can be pre-computed prior to evaluation. Unlike the standard two layer fully connected neural network, our model displays uncertainty for points far away from the training dataset. 
\begin{figure}[t]
  \centering
  \begin{subfigure}{0.49\linewidth}
    \includegraphics[width=\linewidth]{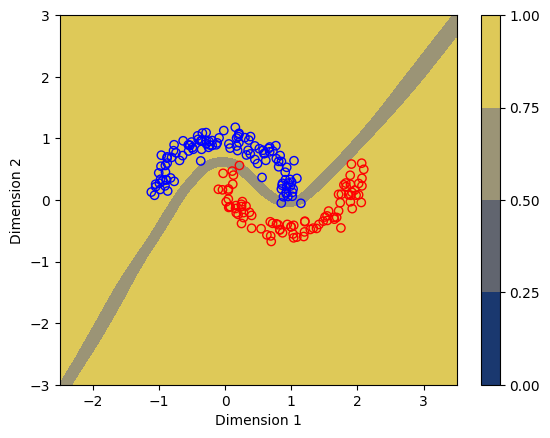}
    \caption{Neural Network}
  \end{subfigure}
  \hfill
  \begin{subfigure}{0.49\linewidth}
    \includegraphics[width=\linewidth]{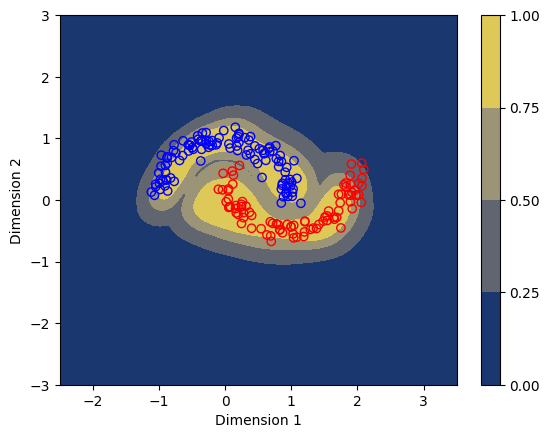}
    \caption{PB\&J}
  \end{subfigure}
  \caption{Out of distribution detection performance on Two Moons dataset. Colorbar represents confidence of prediction.}
  \label{fig:two_moons}
\end{figure}

In OOD experiments with two datasets, we trained the model on the in-distribution (ID) training dataset and evaluated the model on both the ID and OOD test datasets. The same image transformations are applied to both datasets and batch normalization is set to evaluation mode to ensure that the OOD task is not artificially simple \cite{van2020uncertainty}. The area under the receiver operating characteristic (AUROC) is reported as the measure of separability between the ID and OOD datasets. We take the average across five separate runs. For the MNIST vs FashionMNIST task, PB\&J achieves a similar classification accuracy and better AUROC than a standard neural network with the same hyperparameters, as shown in Table 3. For the FashionMNIST vs MNIST task, PB\&J achieves better classification accuracy and better AUROC, which is comparable to DUQ, than a standard neural network, as shown in Table 4. 

\begin{table}[t]
  \centering
  \begin{tabular}{@{}lrrr@{}}
    \toprule
    Method & Accuracy & AUROC \\
    \midrule
    PB\&J & $99.5 \pm 0.03$ & $0.992 \pm 0.001$ \\
    Neural Network & $99.5 \pm 0.05$ & $0.984 \pm 0.004$ \\
    \bottomrule
  \end{tabular}
  \caption{Classification accuracy on MNIST and AUROC with FashionMNIST as OOD.}
  \label{tab:mnist}
\end{table}

\begin{table}[t]
  \centering
  \begin{tabular}{@{}lrrr@{}}
    \toprule
    Method & Accuracy & AUROC \\
    \midrule
    PB\&J & $92.8 \pm 0.09$ & $0.951 \pm 0.006$ \\
    DUQ & $92.4 \pm 0.2$ & $0.955 \pm 0.007$ \\
    Neural Network & $92.4$ & $ 0.843 $ \\
    \bottomrule
  \end{tabular}
  \caption{Classification accuracy on FashionMNIST and AUROC with MNIST as OOD. Results for DUQ and Neural Network are from \protect\cite{van2020uncertainty}. All methods use a three layer convolutional neural network architecture.}
\label{tab:fmnist}
\end{table}

\section{Conclusion} 
We introduce a method for modifying softmax classifiers that both provides explanations of its decision making process and shows promise for OOD detection. Our interpretable-by-design architecture enables us to sample for any set of class examples from the entire training dataset as prototypes to retrieve an instance-based explanation for the model's prediction. Our initial examination of OOD detection performance shows that our method outperforms a standard neural network. Examining our method's capabilities for OOD detection is a promising direction that needs further research. A limitation of our approach is that the prototypes correspond to entire images. Techniques to use parts or portions of the image as prototypes within this framework can be explored in future work. Other future work includes expanding the experiments to scientific datasets and exploring other prototype sampling techniques, such as guided sampling methods to account for past prototypes. The code repository is available at \url{https://github.com/Brown-SciML/pbj}.

\section{Acknowledgments}  
This research was conducted using computational resources and services at the Center for Computation and Visualization, Brown University. This work is partially supported by the SciAI Center, and funded by the Office of Naval Research (ONR) under Grant Number N00014-23-1-2729, and by a Brown University Provost's STEM Postdoctoral Fellowship.

\bibliographystyle{named}
\bibliography{ijcai24}

\newpage
% \onecolumn
\section{Appendix}
\subsection{Architecture and Training Details}
For MNIST and FashionMNIST, we used a three-layer convolutional neural network from \cite{van2020uncertainty} with a latent space dimension of $256$ as our backbone network. We used stochastic gradient descent (SGD) optimizer with an initial learning rate of $0.05$ and a learning rate scheduler that reduced the learning rate by $0.1$ at $25$ and $50$ epochs. The initialization for weight matrix prior to the softmax ($W$) was $\gamma=100$. 

For CIFAR10, we used ResNet18 with a latent space dimension of $256$ as our backbone network. We used SGD optimizer with an initial learning rate of $0.1$ and a cosine annealing learning rate scheduler. The initialization for $W$ was $\gamma=1000$.

For CUB-200-2001, we used ResNet34/ResNet50 with latent space dimension of $400$ as our backbone network. ResNet was pretrained on ImageNet and finetuned for this bird classification task. We used Adam optimizer with a learning rate of 1e-4. The initialization for $W$ was $\gamma=1000$.

\subsection{Explainability with Prototypes Continued}
\label{sec:protocontinued}

We show the posterior distribution and highest scoring prototypes for the first five misclassified test images from the test set for FashionMNIST (in Figure 5) and MNIST (in Figure 6). The first column in Figure 2 corresponds to the example shown in Figure 5. Each test image results in varying degrees of uncertainty between multiple classes. We display the four highest scoring prototypes from the top two most common classes. 

We also show the posterior distribution and highest scoring prototypes for the first five correctly classified test examples from FashionMNIST (in Figure 7) and MNIST (in Figure 8). The first column in Figure 3 corresponds to the example shown in Figure 7. Unlike the misclassified test images, for nine out of ten correctly classified test images, the model is $100\%$ certain of its prediction.

% \subsection{Two Moons}

\begin{figure*}
\centering
\includegraphics[width=0.7\textwidth]{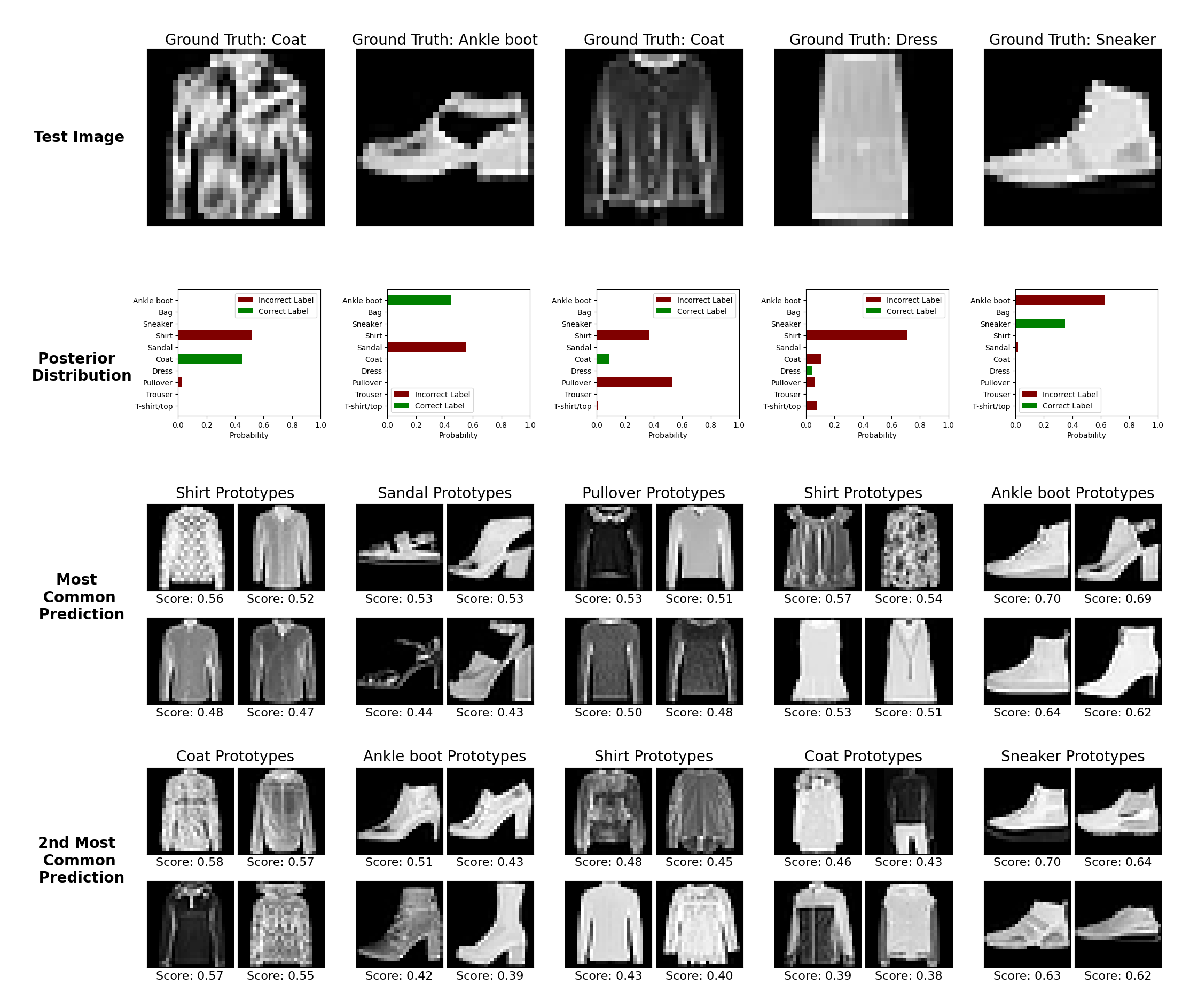}
\caption{Posterior distribution and prototypes for first five misclassified test images from FashionMNIST.}
\label{fig:fmnist_grid_incorrect}
\end{figure*}

\begin{figure*}
\centering
\includegraphics[width=0.7\textwidth]{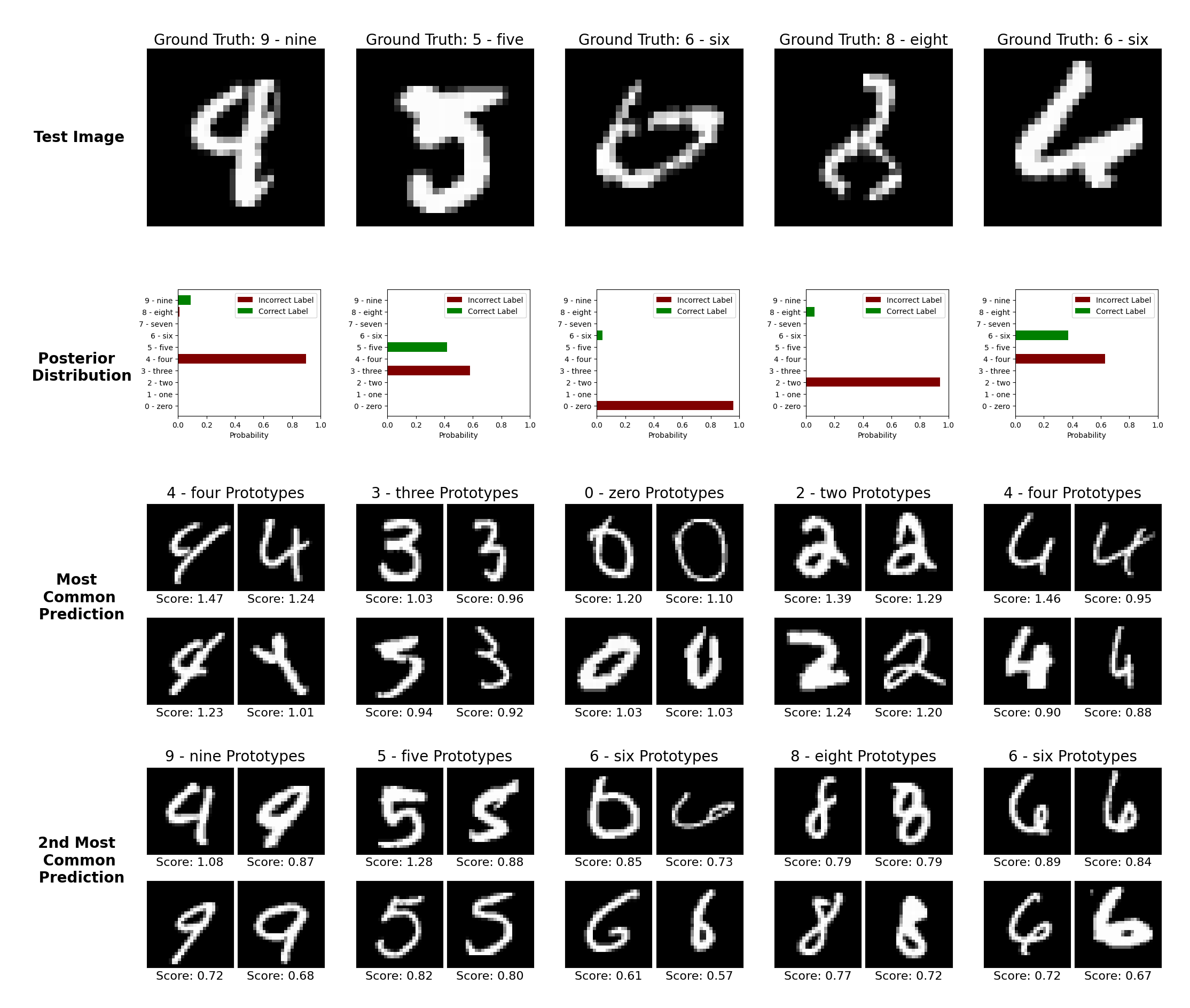}
\caption{Posterior distribution and prototypes for first five misclassified test images from MNIST.}
\label{fig:mnist_grid_incorrect}
\end{figure*}

\begin{figure*}
\centering
\includegraphics[width=0.7\linewidth]{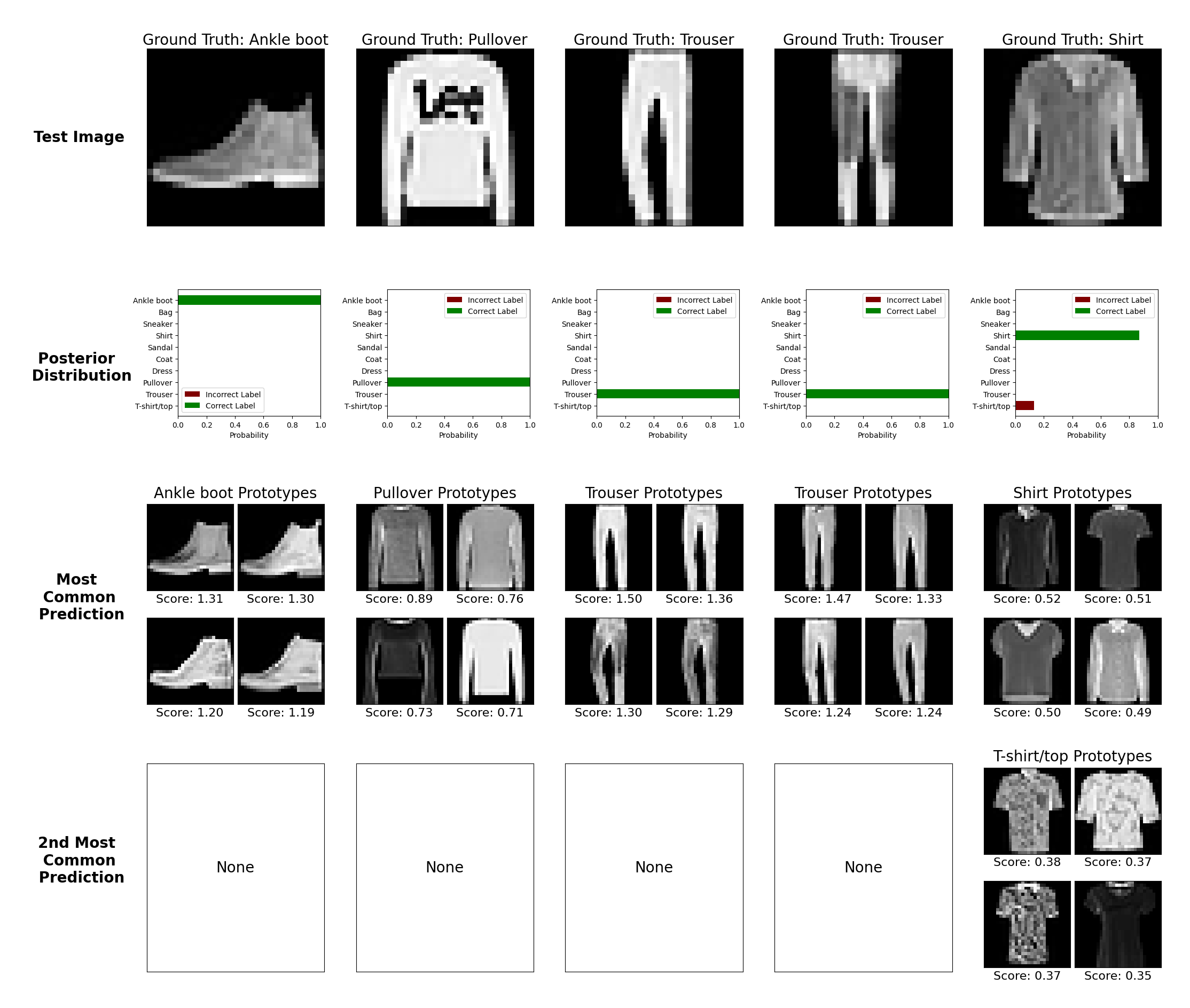}
\caption{Posterior distribution and prototypes for first five correctly classified test images from FashionMNIST.}
\label{fig:fmnist_grid_correct}
\end{figure*}

\begin{figure*}
\centering
\includegraphics[width=0.7\linewidth]{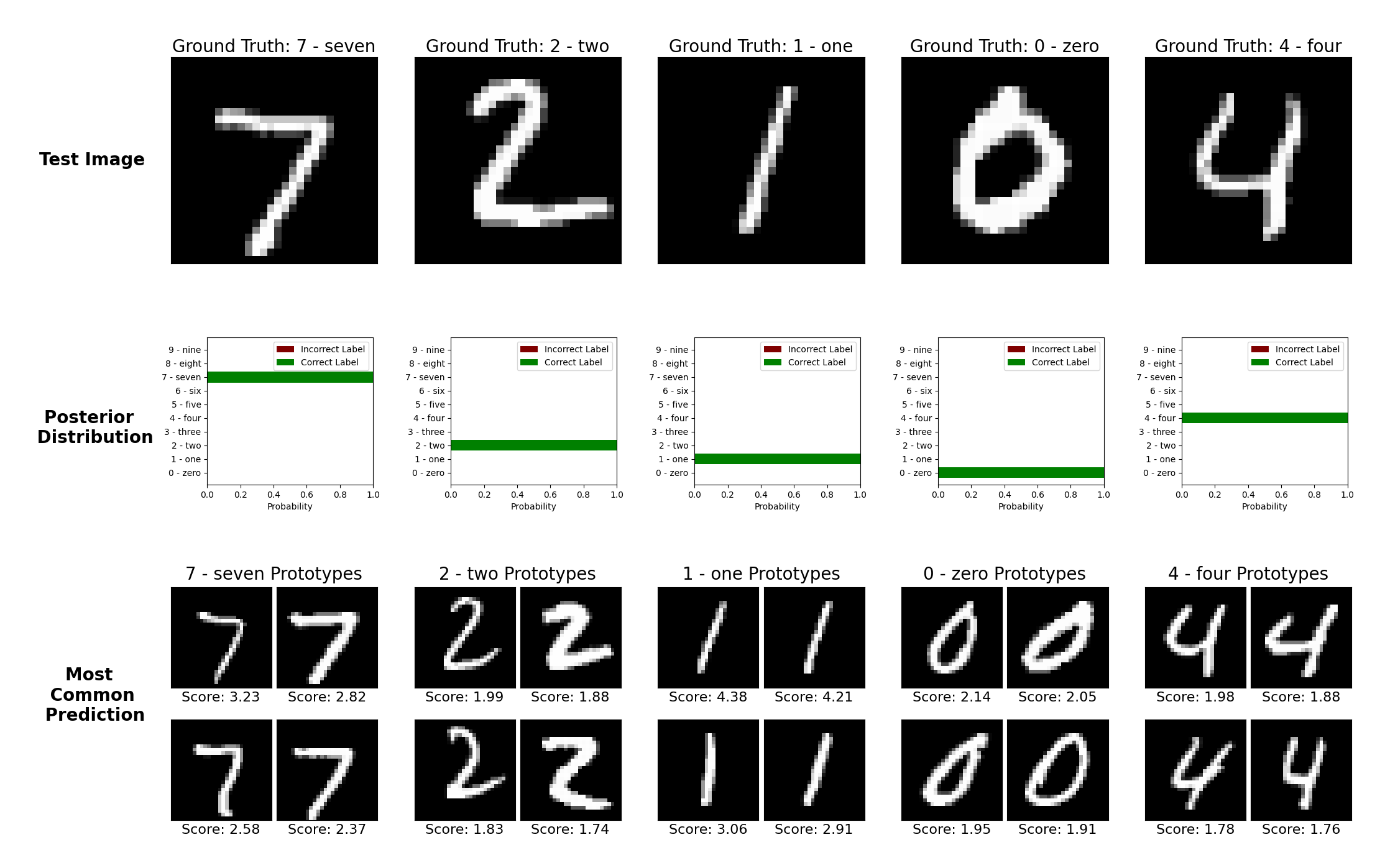}
\caption{Posterior distribution and prototypes for first five correctly classified test images from MNIST.}
\label{fig:mnist_grid_correct}
\end{figure*}

\end{document}